\documentclass{article}

\usepackage{arxiv}

\usepackage[utf8]{inputenc} 
\usepackage[T1]{fontenc}    
\usepackage{hyperref}       
\usepackage{url}            
\usepackage{booktabs}       
\usepackage{amsfonts}       
\usepackage{nicefrac}       
\usepackage{microtype}      
\usepackage{lipsum}
\usepackage{svg}
\usepackage{graphicx}
\graphicspath{ {./images/} }
\usepackage{amsmath} 
\usepackage{pdflscape}
\usepackage{xcolor}
\usepackage[normalem]{ulem}
\usepackage{comment}
\usepackage{longtable}
\usepackage{tabularx}
\usepackage{caption} 
\usepackage{subcaption}
\usepackage{longtable}
\usepackage{booktabs}
\usepackage[utf8]{inputenc}
\usepackage{changepage}

\usepackage{array}
\usepackage{float}
\newlength{\extralength}
\newlength{\fulllength}
\newcolumntype{C}{>{\centering\arraybackslash}X}

\title{Emotion Classification of Children Expressions}
\author{Sanchayan Vivekananthan  \\[1ex]
\begin{minipage}[t]{0.90\textwidth}
\centering
\scriptsize Department of Computer Science, Huddersfield University, Queensgate, Huddersfield HD1 3DH, UK; \\
Correspondence: U2380760@unimail.hud.ac.uk;
\end{minipage}}

\begin{document}

\maketitle
\begin{abstract} 
This paper proposes a process for a classification model for the facial expressions. The proposed process would aid in specific categorisation of children's emotions from 2 emotions namely 'Happy' and 'Sad'. Since the existing emotion recognition systems algorithms primarily train on adult faces, the model developed is achieved by using advanced concepts of models with Squeeze-andExcitation blocks, Convolutional Block Attention modules, and robust data augmentation. Stable Diffusion image synthesis was used for expanding and diversifying the data set generating realistic and various training samples.
The model designed using Batch Normalisation, Dropout, and SE Attention mechanisms for the classification of children's emotions achieved an accuracy rate of 89\% due to these methods improving the precision of emotion recognition in children. The relative importance of this issue is raised in this study with an emphasis on the call for a more specific model in emotion detection systems for the young generation with specific direction on how the young people can be assisted to manage emotions while online.
\end{abstract}


\section{Introduction}
Children in the current world for instance have several opportunities to access any video content online irrespective of its nature being appropriate or otherwise ~\cite{hussain2023child}. Features that in themselves may not be restricted in any way due to other conventional blockage measures such as the keyword blocking or age-blocking will occupy the young people emotionally and psychologically.
More recently, there have emerged technologies such as facial recognition systems that directly explore children and their ability to control content heard or watched in real time. These systems use facial recognition to detect signs of distress, discomfort or fear and prevent the harm by either stopping the video or alerting the caregivers.

One of the measures that will enable the achievement is the formulation of the model that will propose a proper classification of children's facial expressions since these signs differ from adult’s emotional expression both in terms of look and perception. Currently, most of the facial recognition systems are developed based on adult faces, therefore high percentage of wrong results are expected for children’s faces. Therefore, there is a need to develop a specialized model focused on identification and categorization of children’s emotions.

In this paper, some important facial feature differences between children and adults are described and child facial expression recognition model is presented. It offers the review of the literature regarding current models that were derived from adult based and child focused approaches, along with their strengths and weaknesses. The presented methodology describes the design of a new CNN to distinguish between the two considered expressions “HAPPY” and “SAD” with high efficiency. Eight experiments are performed to analyse the model’s performance by varying the regularisation and the type of attention. Evaluation methods involves key indicators like accuracy, training and validation loss, confusion matrices, as well models’ interpretability by Grad-CAM visualizations.

\subsection{Need for children’s expression focused model}
The facial expression patterns significantly differ with different age groups. Howard et al \cite{nref1}{} state that there are clear expression differences between young people and older people. These variations create some difficulties for current emotion recognition models, where many of them are trained on datasets that contain a vast number of adult faces. This age-related gap contributes a lot in lowering the model's accuracy. In support of this, Sayin and Aksoy \cite{nref2} point out a significant gap in research on the classification of facial expressions in younger children and older adults, It might be challenging to use traditional emotion identification techniques meant for adults on children because of their very diverse emotional manifestations resulting from developmental changes, mental progress, and individual character development.

Among some studies, Guodong Guo et al \cite{nref3} focused on the effects of aging on FER, indicating the drastic difference between children and adults. The researchers also mentioned that children mainly communicate by making rigid, clear movements of their face compared to adult people who demonstrate subtle, tonned movements of muscles because of the elasticity of the skin and presence of wrinkles. Likewise, Houstis and Kiliaridis \cite{nref4} pointed out that adult especially male show greater amount of vertical movement in facial expressions, especially when compared with children who have immature muscle hence resulting to differences in feature in expressions between different age groups. In support of this argument, Guo brings a cultural perspective into the equation and argues that the variations must be taken into consideration to enhance FER systems’ effectiveness throughout the developmental period.

Rao et al \cite{nref5} compared the expression detection in both children and adults, and found that the children’s expression was always weaker and not very distinguishable as compared to adults, which reflected the lower accuracy in the recognition models irrespective of whether the less number of facial landmarks were used. In addition to this, another influence to emotional display is cultural and ethnic practices. For example, Lewis et al \cite{nref6} found that Japanese children report lower levels of expression for shame, pride and sadness but higher levels for embarrassment than American children. Camras et al \cite{nref7} also noted that Chinese babies are less expressive than European American and Japanese babies while smiling and crying. These studies demonstrate that cultural aspects are important in emotional display, while further complicating the models of emotion recognition.

\section{Literature Review}
The Mean Supervised Deep Boltzmann Machine (msDBM) is a relatively unexplored technique that Nagpal et al \cite{lref1} introduce as a solution to the problem of facial expression classification in youngsters. msDBM outperformed previous models in tests conducted on the Radboud Faces and CAFE datasets, obtaining accuracy rates of 48.0\% on CAFE and 75.0\% on Radboud Faces. Its usefulness in recognising children's facial expressions is highlighted by its superior performance when compared to traditional and deep learning methods. The study emphasises the need for specialised methods in this area and urges more investigation to handle the difficulties and improve the precision of classifying child expressions.

Howard et al \cite{lref2}  analyse particular issues of recognising biases of the emotional recognition in the context of cloud services that are biased to adults. They introduce a hierarchical classification approach aimed at improving accuracy for underrepresented groups, such as children and specific ethnicities, by integrating pre-trained models with specialized learners for these populations. By this method they enhaced the overall model accuracy by 17.3 \% and  acheived 41.5\% increment in accuracy for the most complex minority images. The study highlights the importance of adapting general models to account for diverse populations and advocates for continued research to enhance bias mitigation and optimize recognition systems across different demographic groups.

Lopez-Rincon \cite{lref3}  conducted a study on emotion recognition in children using the NAO robot, comparing the performance of the AFFDEX SDK and a Convolutional Neural Network (CNN) combined with Viola-Jones, which was trained on AffectNet and fine-tuned using the NIMHChEF datasets. AFFDEX properly recognised 506 out of 1,179 faces, whereas CNN correctly identified 535 out of 1,192 faces, demonstrating greater overall accuracy. Unfortunately, CNN struggled to identify disgust, most likely because there weren't many cases in the dataset. The accuracy rose to 46.05\% when both systems were combined into an ensemble, indicating the possibility for improved real-time emotion recognition with more inputs and techniques.

Yu \cite{lref4}  uses a combination of actions, facial expressions, and contextual data to investigate how preschoolers recognise emotions. In order to address data imbalance, hierarchical sampling is used in the study to improve the efficiency of the deep learning model, and a novel feature descriptor for facial variations is introduced. Moreover, the method incorporates audio and geometric aspects to improve accuracy. The model performed well on the BP4D and MMI datasets and attained 97.8\% accuracy on the CK+ database. The results highlight how well appearance and geometric features work together with LSTM models to enable multi-modal emotion recognition in real-time under dynamic conditions.

To tackle with this issue, Sayin and Aksoy \cite{lref5}, devoted their work to detecting expressions in children, which is one of the groups not considered frequently in such experiments. By gathering pictures of children's faces from search engines and classifying them into seven categories which are, anger, disgust, happiness, neutrality, sadness, fear, and surprise. Six deep learning models namely VGG 16, ResNet 50, DenseNet 121, Inception v3, Inception ResNet v2 and Xception models were applied using transfer learning to assess the performance. With an F1 score of 0.76, InceptionV3 attained the maximum accuracy of 76.3\%. With "happy" faces, all the models did well; nevertheless, they had trouble with "scared," "surprised," and "sad" expressions. In order to avoid bias in smart systems, the study emphasises the importance of including a range of age groups in facial expression research.

In a study by Witherow et al \cite{lref6}, the authors investigate the application of transfer learning for classifying facial expressions in children. They first train a convolutional neural network (CNN) using adult photos from the CK+ dataset, then fine-tune it using the CAFE dataset, in an attempt to overcome the difficulty presented by the lack of child-specific datasets. Five distinct model configurations were tested in their trials, and the accuracy rates obtained with the CK+ dataset were 93\%, the CAFE dataset alone was 63\%, and the CK+ model was 76\% accurate when it was adjusted with the use of that data. Using the fine-tuned model, the best results were obtained on the CAFE dataset, indicating that transfer learning can improve the classification of child expression. This research shows that models trained in adult data can be effectively transfer to children, thus raising questions on the efficiency of using only adult data. The authors are also planning on extending this approach by incorporating more comprehensive data and various model structures to enhance the outcomes of child facial expression recognition and facilitate the early detection of developmental disorders, including autism spectrum disorder (ASD).

Banerjee et al \cite{lref7} conducted research aimed at enhancing facial expression recognition models for mobile devices, targeting individuals such as children with autism who face challenges in recognizing emotions. The research used MobileNetV3-Large and other convolutional neural networks, as well as trained models on datasets with images of children, images of adults, and images of both children and adults together. The experiment was done on a Motorola Moto G6 model note which acts as the system for every model. After training on all available data, MobileNetV3-Large demonstrated minimum latency and attained an accuracy of 65.78\% and an F1-score of 65.31\% on the Child Affective Facial Expression dataset. The outcomes showed that models, which were trained on data from children’s database, had better performance and that there were disparities between ethnic groups results which emphasized the necessity to increase the models’ effectiveness in terms of underrepresented ethnicities.

Talaat \cite{lref8} designed a real-time emotion recognition system tailored for autistic children, targeting the detection of six emotions: anger, fear, joy, neutral, sadness and surprise.  For feature extraction and selection, the system makes use of a deep convolutional neural network (DCNN) framework with an autoencoder. Tests were conducted using pre-trained models, including ResNet, MobileNet, and Xception; Xception produced the best results. Its accuracy was 95.23\%, its sensitivity was 0.932, its specificity was 0.9421, and its AUC was 0.9134\%. The system can also combine the fog and IoT technologies with higher efficacy.

Rani \cite{lref9} examined the identification of four emotions on the face of autistic children: neutrality, rage, sadness, and happiness.  Local binary patterns (LBP) were used for feature extraction in the study's image processing and machine learning methods, which included support vector machines (SVM) and neural networks. According to the results, the neural network with LBP only attained 70\% accuracy, while the SVM with LBP achieved 90\% accuracy. The study emphasises how useful SVM is in this situation and makes recommendations for future developments that could increase the suggested system's ability to identify emotions in autistic youngsters.

Begeer et al \cite{lref10} investigated the emotional expression responses of high-functioning autistic children. Participants in the study, which consisted of 31 boys with typical development and 28 boys with autism, were given the job of classifying pictures of adults with frowning or happy expressions. The results demonstrated that, in neutral environments, autistic children paid less attention to emotional cues, but when faced with socially significant decision-making, they matched the attention spans of their typically developing peers. This suggests that situational stimuli may improve the emotional understanding of autistic children by influencing their emotional attention. The study focusses on how task demands can affect how autistic youngsters interpret their emotions.

\section{Available Child Facial Expression Datasets}
The creation of reliable emotion classification models for children heavily depends on the availability of high-quality, labeled datasets. However, there are some other large-scale facial expressions database such as CIFAR-10 dataset, AffectNet, FER+, and the Oulu-CASIA NIR and VIS facial expression database \cite{aref1}, most of them include mainly adult faces and have limited information regarding children facial emotions.

On the other hand, there are several datasets that have been developed to capture children’s facial expression exclusively, some of which are; The Children Affective Facial Expression (CAFE) Database \cite{aref2}, City Infant Faces Database \cite{aref3}, Children Spontaneous Facial Expression Video Database (LIRIS-CSE) \cite{aref4}, Dartmouth Database \cite{aref5} of Children's Faces, and ChildEFES Dataset \cite{aref6}.

\begin{itemize}
\item \textbf{CAFE Database:} The CAFE database is one of the limited publicly available resources focused specifically on children's facial expressions. Pictures of children between the ages of two and eight are included, along with labels for various emotions like fear, rage, grief, and joy. Still, its value for training deep learning models is limited because of its relatively small size when compared to adult-focused datasets. To further develop emotion detection models that generalise well across many populations, the dataset lacks diversity in terms of age range, ethnicity, and cultural representation.

\item  \textbf{City Infant Faces Database:} This dataset involves babies and it covers almost all facial expressions in various scenarios.

\item \textbf{Children Spontaneous Facial Expression Video Database (LIRIS-CSE):} This dataset includes video data of the children’s natural and voluntary facial expressions in response to stimuli, which can be considered an advantage over the previously mentioned static ones.

\item \textbf{ChildEFES Dataset:} It is a database of children’s facial images and videos, which has been developed specifically for the purpose of emotion recognition tasks.

\item \textbf{Dartmouth Database of Children's Faces:} This dataset is particularly beneficial for studying children’s facial expressions as it contains the pictures of 80 Caucasian kids, from 6 to 16 years old, which performed 8 different emotions in various lighting and angle settings.

\end{itemize}

Despite the potential of such datasets, the use of even specialised datasets can be hampered by issues such as data set size, availability and diversity. These problems can, however, be solved by image synthesis. Creating synthetic facial images of children in a given emotional state will improve the performance of emotion classification models since datasets can be designed to overcome current limitations.

\section{Role of Synthetic Data in Enhancing Datasets}
In machine learning, the quality of the dataset is crucial for training efficient models ~\cite{aydin2023domain}, particularly in projects where the goal is to identify emotions in children's facial expressions. In addition to being sufficiently large to encompass a broad spectrum of emotions, a useful dataset should accurately depict the situations that the model is intended to analyse. A dataset that has been carefully chosen makes it possible for the model to more fully represent the underlying patterns in the data, which improves the model's ability to generalise to new, unforeseen cases, not only for child emotion but also other applications such as Manufacturing ~\cite{hussain2023yolo} and Healthcare ~\cite{hussain2022exudate}.

According to Tanaka and Aranha \cite{rref1}, the creation of synthetic data is important for dataset augmentation since it may be used for several objectives, including protecting privacy and oversampling minority classes. In tasks such as emotion recognition, where specific expressions might not be as prevalent, this method can prove to be quite advantageous.

\begin{enumerate}
    \item Oversampling Minority Classes : It is not uncommon to have the class distribution skewed in datasets used for emotion detection with “happy” expressions far outnumbering “sad” ones. This imbalance results in models that are more likely to output “happy” and fail in the production of a correct “sad” label. To fix this challenge, synthetic data can be created to supplement the minority class. Substituting some parameters in the original examples generates new instances of emotions getting closer to the target population, which enhances the model’s generalisation and diversity.

    \item Preserving Privacy with Synthetic Data: One of the challenges faced when using such information involves privacy issues such as working with children’s facial expressions. Synthetic data can be defined as a set of completely fake data still maintains statistically similar properties to the original data but with obscured identifiable features. Learned patterns of producing realistic and varied synthetic images can be involved Stable Diffusion, Variational Autoencoders (VAEs), Generative Adversarial Networks (GANs), and SMOTE. All the techniques used here preserve the privacy of those involved, ensure model training success, and conform to the law and ethics.

\end{enumerate}

\section{Methodology}
\subsection{Image Collection}
To establish a dataset suitable for training and testing the classifier for child facial expressions, the internet was searched for child photographs with no copyright. The focus was on two primary emotions: Happy and Sad. The images collected were 180 in number where 100 images recorded Happy and 80 images recorded Sad. To make the classification process effective, a review of the classification was made by two random individuals to perform the secondary rating of each image re-categorising the images based on the most predominant emotional state detected. This step was important to ensure that the images respectively belonged to their each classes. A few example images from the dataset are shown in Figure \ref{Figure:1} below.

\begin{figure}[H]
\begin{adjustwidth}{-\extralength}{0cm}
\centering
\includegraphics[height=5cm]{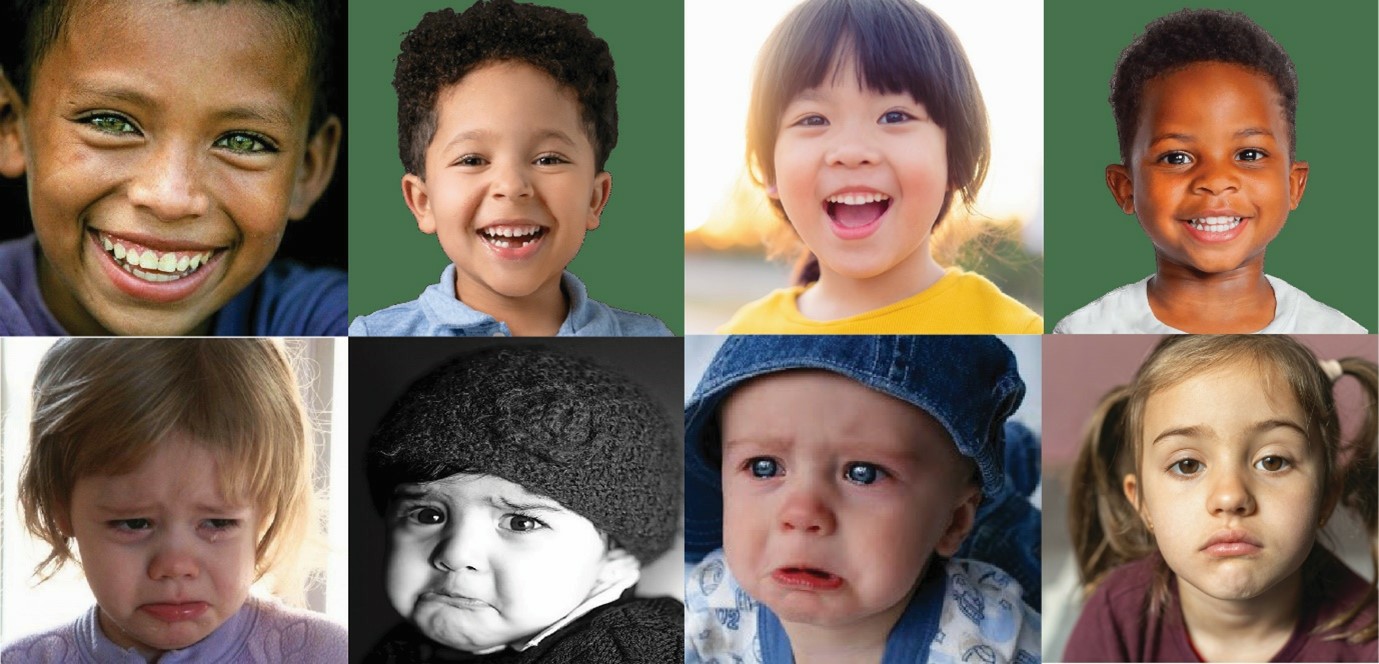}
\end{adjustwidth}
\caption{Sample images from the dataset collected from the internet}
\label{Figure:1}
\end{figure}

\subsection{Image Synthesis}
To address the limited number of image samples especially for Sad expression, several image synthesis approaches were employed to expand and enrich the dataset. Four methods were explored: This would include Generative Adversarial Networks (GANs), Variational Autoencoders (VAEs), Stable Diffusion, Grounding DINO + SAM, and Stable Diffusion. The overall structure and functionality of each of the examined techniques was discussed in an effort to figure out how they contribute to the creation of synthetic images. The methods were then tested based on the level of sophistication with which facial expressions appeared realistic and diverse to enhance the quality of the dataset for classification.

\paragraph{\textbf{Generative Adversarial Network (GAN)}}
\mbox{}\\ 
\mbox{}\\
Generative Adversarial Networks (GANs), introduced by Goodfellow in 2014, consist of two neural networks: a Generator and a Discriminator. The Generator creates fake data and the Discriminator is to determine whether the data is real or fake data. Both compete with each other, where the Generator adds its effort to generate data that will not be discriminated upon by the Discriminator, and on the other side, the Discriminator works on enhancing its ability to detect false data. This architecture makes GANs capable of synthesizing new realistic data especially in image generation. Figure \ref{Figure:2}  below illustrates the architectural representation of the GAN model.

\begin{figure}[H]
\begin{adjustwidth}{-\extralength}{0cm}
\centering
\includegraphics[height=6cm]{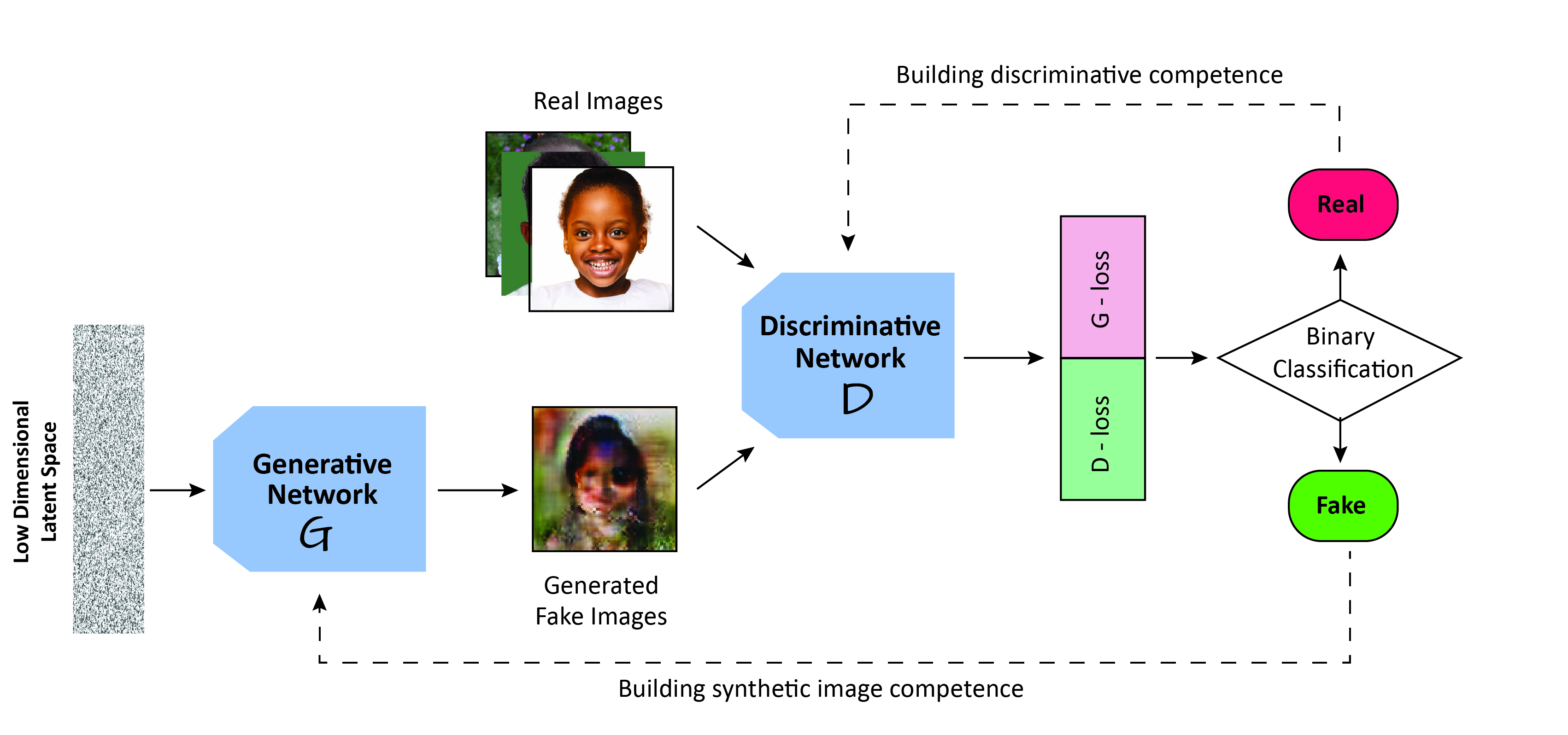}
\end{adjustwidth}
\caption{Generative Adversarial Network (GAN) Architecture.
Reprinted from Vivekananthan (2024).
}
\label{Figure:2}
\end{figure}

GANs come in a variety of forms, each designed for a particular use case. These include Vanilla, Conditional \cite{Gref1}, Deep Convolutional \cite{gref7}, CycleGANs \cite{gref8}, StyleGANs \cite{gref9}, and Super Resolution GANs \cite{gref10}. Due to its superior picture generation capabilities over models such as Variational Autoencoders (VAEs) in terms of flexibility and realism, GANs have gained popularity. They do, however, have certain drawbacks, such as unstable training, mode collapse, expensive computing, and the possibility of overfitting, which can reduce the variety of images. Furthermore, the assessment of outputs generated by GANs is still subjective and lacks dependable measures.

\begin{figure}[H]
\begin{adjustwidth}{-\extralength}{0cm}
\centering
\includegraphics[height=6cm]{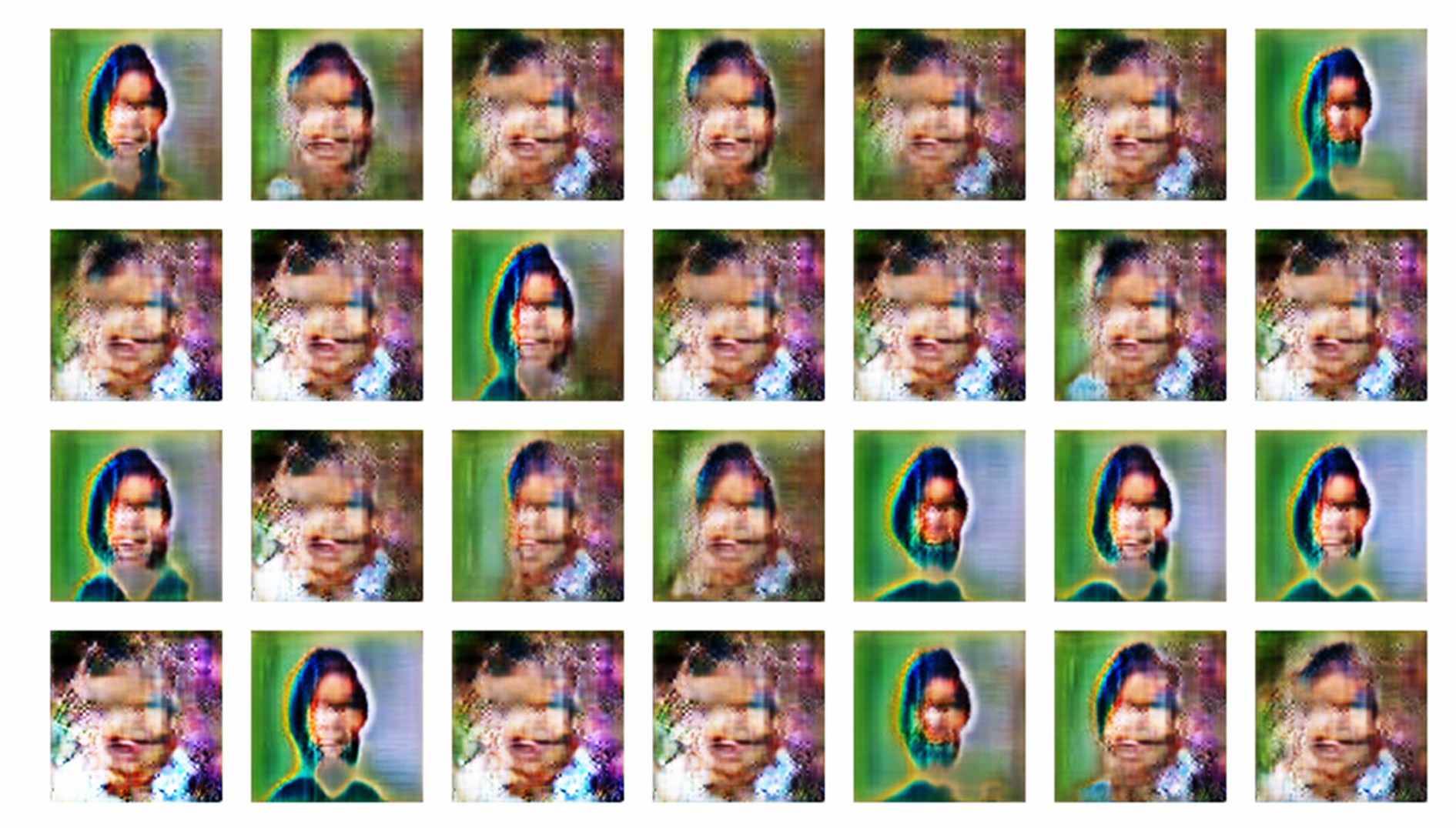}
\end{adjustwidth}
\caption{Samples of the synthesized images using GAN technique}
\label{Figure:3}
\end{figure}

In our project, Generative Adversarial Networks (GANs) were used to create more images in an attempt to augment the set. However, because of the resource and time constraints in training the model, the epoch was limited to 250 only. As depicted in figure \ref{Figure:3} above, it was observed that the resulting images were below the required image quality. Still, many were out of focus, slightly blurred and often devoid of detailed faces, which limited their practical application for the work. 

This outcome demonstrates some of the challenges of deploying GANs under such constraints. The problems with the image quality mentioned in the paper illustrate that generating high-quality synthetic images, in general, is a process that does not take less time and tends to demand more computational resources. It is therefore evident that the difficulties encountered in this project reflect just how much training duration and available resources affect the efficiency of GAN-generated data.

\begin{figure}[H]
\begin{adjustwidth}{-\extralength}{0cm}
\centering
\includegraphics[height=3.5cm]{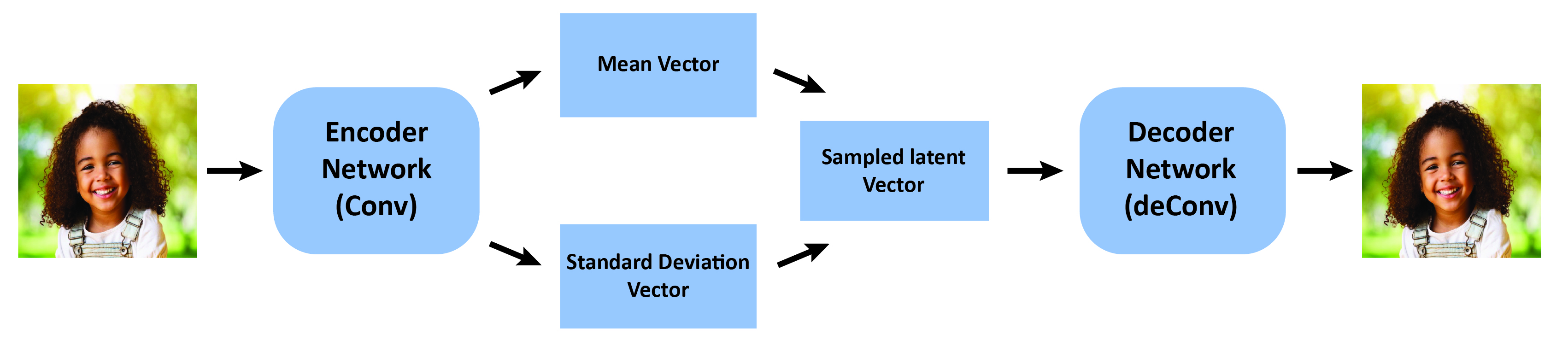}
\end{adjustwidth}
\caption{Variational Auto Encoder (VAE) Architecture.Reprinted from Vivekananthan (2024)
}
\label{Figure:4}
\end{figure}

\paragraph{\textbf{Variational Auto Encoder (VAE)}}
\mbox{}\\ 
\mbox{}\\
Variational Autoencoders (VAEs) are deep, generative models which are intended for learning implicit distributions and are a combination of autoencoders and probabilistic approaches that produce data that is similar to the training dataset \cite{ref2}. As per the figure \ref{Figure:4} a VAE consists of two main components: an encoder and a decoder \cite{Gref5}.  The encoder turns the input data into the low dimension space by computing parameters such as mean and variance of the Gaussian function. The decoder then reconstructs the original data from this latent code it is given.  Depending on the data type, the fully connected or convolutional structures could be applied for both the encoder and decoder.

Training denotes the process of finalising the model via stochastic gradient descent and the re-parameterization technique \cite{ref4}. VAEs are shown to be capable of modelling complex distributions and work well for large datasets but have issues with retaining the fine architectures of the images and reduced sharpness of the reconstructions because of the loss functions used. Also, problems such as posterior collapse may arise, where the network fails to consider the latent variables hence fewer diverse outputs are generated \cite{Gref5}.

In our project, Variational Autoencoders (VAEs) were used for data augmentation hence creating synthetic images for the dataset. The VAE model was trained over 250 epochs, indicating a much faster training profile compared with many image generation methods. However, they were not able to produce images of good quality despite the efficiency of the process.

\begin{figure}[H]
\begin{adjustwidth}{-\extralength}{0cm}
\centering
\includegraphics[height=7cm]{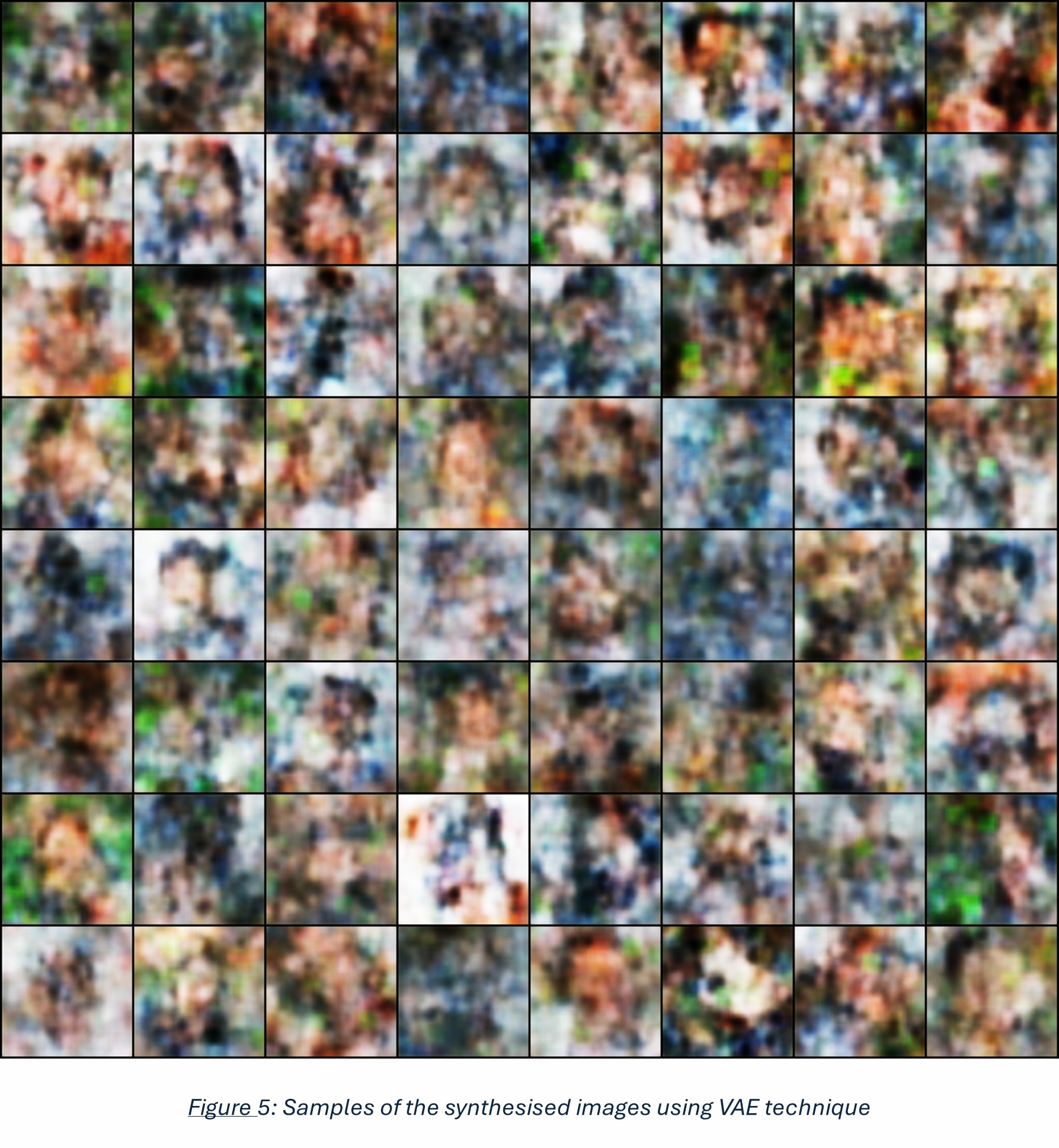}
\end{adjustwidth}
\caption{Samples of the synthesised images using VAE technique
}
\label{Figure:5}
\end{figure}

As it has been illustrated in the Figure \ref{Figure:5}, the images reconstructed by the VAE model were almost blurry and sometimes failed to visualise even primary distinguishable features of a face. This outcome illustrates a common limitation of VAEs which is theinability to fine-tune and produces images of high quality. Time duration and the computational power available for the successful completion of the project restricted other attempts to fine tune the generated images using VAE.

\paragraph{\textbf{Stable diffusion}}
\mbox{}\\ 
\mbox{}\\
Nevertheless, GAN has its downs such as mode collapse and adversarial domain and problem with VAE of having blurry outcomes while Stabled Diffusion comes out a clear winner compared to the above outlined models because of their inability to generate high-resoluted detailed and diverse pictures. Diffusion models have found application in various generative tasks, including image generation \cite{sref1,sref2}, image-to-image translation \cite{sref3}, image editing \cite{sref5}, image inpainting \cite{sref7}, and image super-resolution \cite{sref9}. Discriminative tasks like segmenting and classifying images are also performed by these models.

\begin{figure}[H]
\begin{adjustwidth}{-\extralength}{0cm}
\centering
\includegraphics[height=5cm]{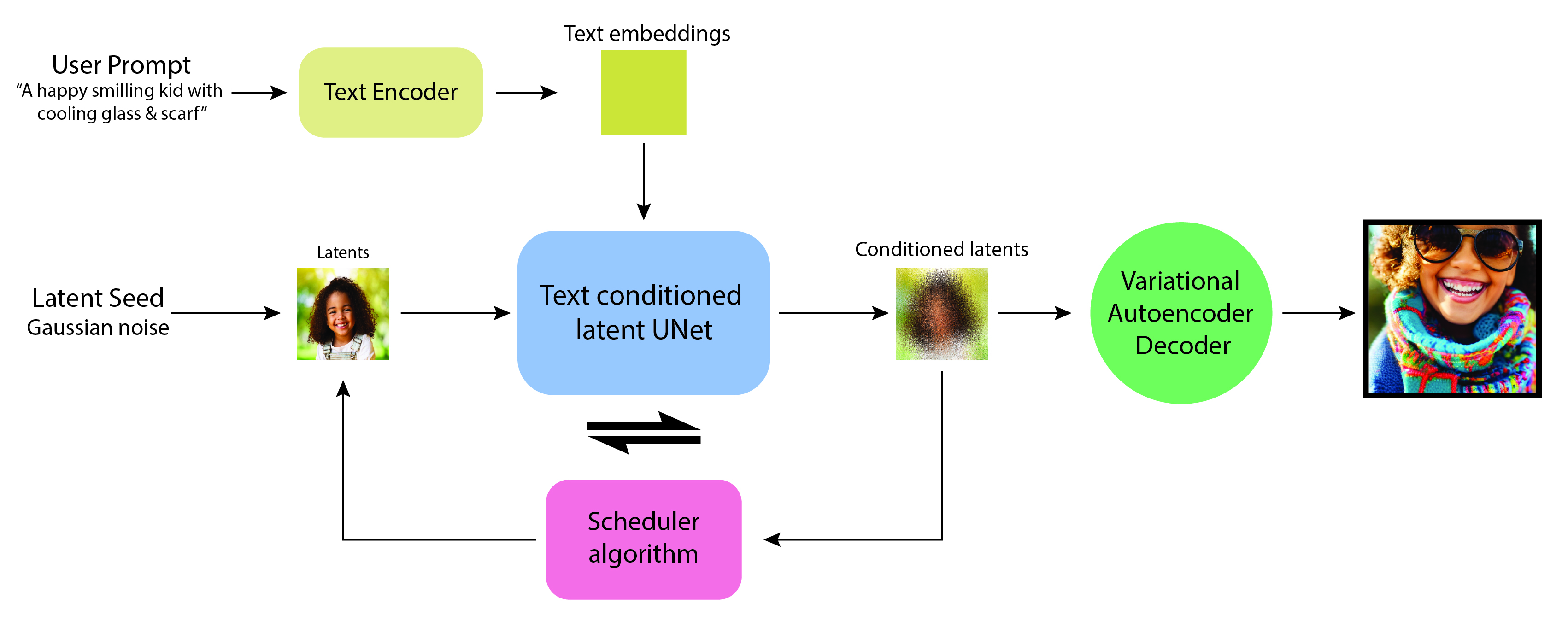}
\end{adjustwidth}
\caption{Stable Diffusion Architecture. Reprinted from Vivekananthan (2024)}
\label{Figure:6}
\end{figure}

The generation process is based on a reverse algorithm involving three key components: a VAE, a U-Net, and a text encoder. As shown in figure \ref{Figure:6}  initially, a VAE compresses images into latent codes, reducing computational demands. The model undergoes two stages which are forward diffusion, where noise is added to the latent representation, and reverse denoising, where a U-Net predicts and removes the noise to reconstruct the image. The ResNet layers and Vision Stack applied in the U-Net have improved image quality. For instance, to support the U-Net to map between images and textual descriptions, a textual encoder including CLIP encodes the input prompts into vectors. Finally, the VAE decoder is used to generate high-resolution images from the learned latent representation \cite{sref11}.

While using Stable Diffusion for image synthesis, the algorithm shows high quality of synthesised images with semantic continuity and richness of varied topics, at the same time it lacks efficiency in terms of time. The generation process involves numerous steps of evaluation in order to complete a single image, and such an approach is slow for practical use in cases where image production should be fast.

\paragraph{\textbf{Advanced Image Synthesis with Grounded DINO, SAM and Stable Diffusion in-painting}}
\mbox{}\\ 
\mbox{}\\
Among the state-of-art methods in the area of image synthesis, Stable Diffusion has been identified as generating sharp images with fine details. However, combining technologies like Grounding DINO and Grounded SAM even adds more value to them. By grounding DINO, the masks are refined for object detection and image context by making sure inpainting is aligned with content and structure of the image. Grounded SAM helps in training the model to generate accurate boundaries of the segmentation masks for certain regions in order to inpaint. Inpainting, combined with Stable Diffusion, improves image details by relying on context to fine-tune inpainting. 
\begin{figure}[H]
\begin{adjustwidth}{-\extralength}{0cm}
\centering
\includegraphics[height=6cm]{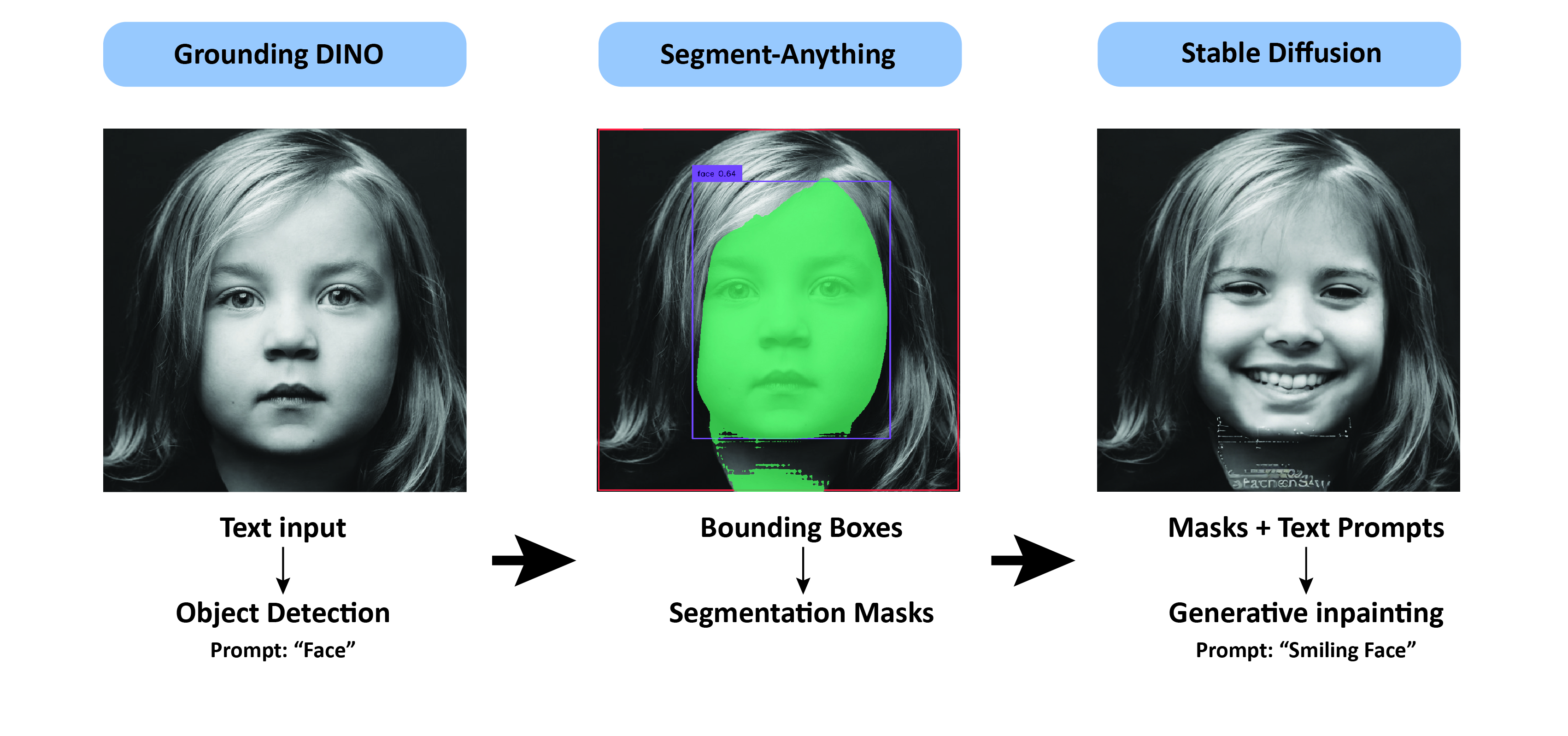}
\end{adjustwidth}
\caption{Advanced Image Synthesis with Grounded DINO, SAM and Stable Diffusion inpainting. Reprinted from Vivekananthan (2024) }
\label{Figure:7}
\end{figure}

provides several benefits, including as increased contextual awareness and precision through precise segmentation, as illustrated in Figure \ref{Figure:7} . Although with such approach, the outcomes are more precise and contain more comprehensive information, this approach calls for many calculations, high levels of specialisation, and more time to complete the process. Also, the usage of certain techniques may result in overfitting, and, thus, restrict the applicability of the method. Another problem is identifying and changing the fine features, which may not be easily recognised by prompts.

It was chosen to use the method named Stable Diffusion for image synthesis to enrich the dataset. This was because it created images that where more accurate and clear as compared to the other methods. The faces generated by Stable Diffusion not only resembled realistic human faces but also were able to express emotions needed for proper emotion identification. Also, for large images, it provided the advantage of being less expensive and taking shorter synthesis time. By adopting Smooth Diffusion approach, 500 images of Happy Expression and 520 images of Sad Expression were generated.

\begin{table}[H]
\centering
\caption{Original and Synthesised Images in the Dataset}
\label{table:dataset 1}
\begin{tabular}{|p{3.5cm}|p{3.5cm}|p{4cm}|p{4cm}|}
\hline
\textbf{Class} & \textbf{Original Images} & \textbf{Synthesised Images}\\ \hline
\textbf{Happy} & 100 & 500\\ \hline
\textbf{Sad} & 80 & 520\\ \hline
\end{tabular}
\end{table}

In combination with SAM and Stable Diffusion, they used grounding of DINO in addition to Stable Diffusion to enhance expression and occlusion. Stating DINO and SAM was beneficial to control occlusions and keep face features in the synthesised images. Moreover, by increasing smiles for ‘Happy’ and enhancing the facial gesture for ‘Sad,’ these strategies increased the utilitarian augmentation since it provided a range of emotional results.

\begin{figure}[H]
\begin{adjustwidth}{-\extralength}{0cm}
\centering
\includegraphics[height=6cm]{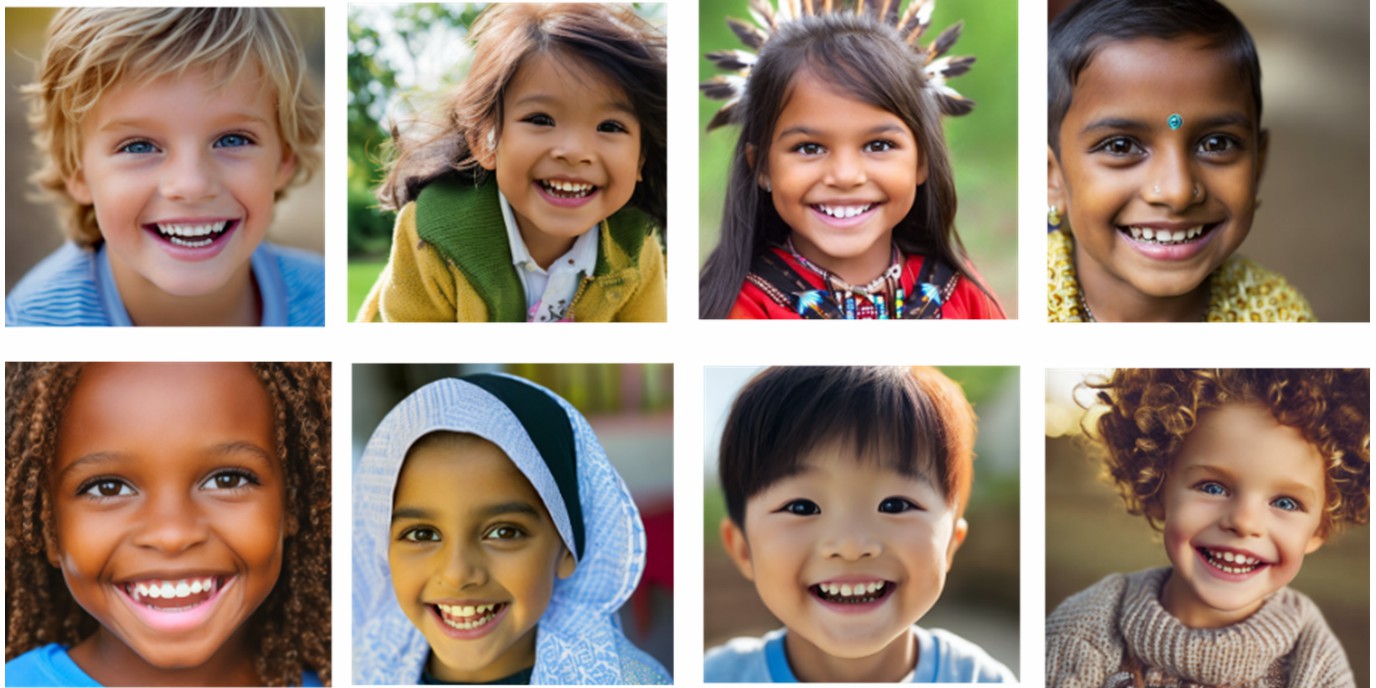}
\end{adjustwidth}
\caption{Synthesized Image Samples Representing Diverse Ethnic Groups. }
\label{Figure:8}
\end{figure}

A number of functional objectives that had been mentioned in the literature were prioritised heavily during the picture synthesis process. The task of enriching the dataset with American Indians, Asians, Black or African Americans, Native Hawaiians, Hispanics or Latinos, and Whites involved the inclusion of a variety of ethnic groups. The backgrounds were also changed to reflect the mindsets and environments seen in these places. A sample of the synthesised images that represent diversity is seen in figure \ref{Figure:8} , which is provided below.

Besides increasing the number of phrases for diversity, the synthesis also aimed at increasing the overall size of the dataset, the precision in measuring the emotions, and the variability in context. Extra care was taken to align the synthetic images in such a way that the emotions of the expressions would still be captured despite having occlusions such as hair across the face, sunglasses, scarves, or even hats. The synthesised occluded pictures are shown in figure \ref{Figure:9}  which is given below.

\begin{figure}[H]
\begin{adjustwidth}{-\extralength}{0cm}
\centering
\includegraphics[height=3cm]{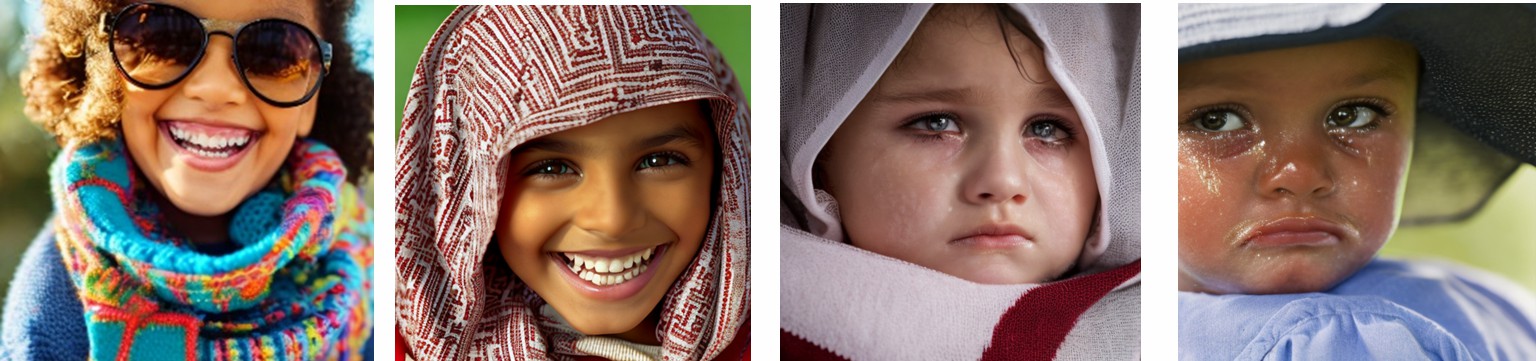}
\end{adjustwidth}
\caption{Samples of Synthesized images with Occlusion }
\label{Figure:9}
\end{figure}

To evaluate the model adequately, the dataset used was partitioned into training and validation set. Thus, out of 1200 photos, 480 were happy and 480 were sad photos in the training set, according to the table above 240 photos included 120 happy and 120 sad photos which constituted of 20\% of the main dataset. The model is sufficiently tested on the exemples that were not included in the construction of the training and the validation set. The model has enough data it can learn from due to a balanced distribution of the two classes in the set. This results into a more reliable and accurate classification performance as compared to the other approaches.

\subsection{Data Augmentation}
After generating additional training images through data synthesis, data augmentation became crucial for preparing the dataset for emotion classification. While synthesis established a foundational image set, augmentation was key to enhancing the dataset's variability and robustness. By applying various transformations, augmentation simulated real-world variations, enabling the model to better generalize to unseen data.

The augmentation of data is crucial for the detection and categorisation of emotions, as shifts in focus, lighting, and angles which are frequently present in facial expressions can change the essential features. In addition, the size of the dataset is artificially inflated through data augmentation while some augmentations are motivated by more specific reasons that are likely to occur in real-world applications. This will further improve abilities of the model to recognise and characterise the emotions as well as reduce the possibility of overfitting in a variety of contexts.

\subsubsection{Need for Data Augmentation}
\begin{itemize}
    \item Expand the Size of the Dataset: Typically, datasets for emotion detection are tiny and unbalanced. By producing more samples, augmentation lessens overfitting and enhances generalisation.
    \item Increased Robustness of the Model: Variability introduced by augmented data helps the model adapt to variations in illumination, perspective, expressions on the face, and other factors.
    \item Lessen Overfitting: By giving the model a wider range of instances, overfitting is lessened because the model learns more effectively and doesn't retain the training set. 
    \item Enhance Model Performance: By giving the model a wider range of training instances, augmented data can improve performance measures.
\end{itemize}

\subsubsection{Data Augmentation methods for emotion detection}
To ensure the model could handle diverse real-world challenges in emotion detection, three main types of augmentations were applied: The first type of variability is Image Appearance Variations, Geometric and Pose Variations, Occlusion and Noise Variations. These augmentations were intended to replicate real-world scenarios to improve the model’s robustness.

Image appearance variation: Image Augmentation Techniques such as contrast, brightness, and colour added to the model enhances the ability to tackle issues of lighting and image quality. Changes were made concerning contrast and brightness and also Concerning colour jittering there were changes that affected hue, saturation and brightness in that; Other pre-processing techniques included Gaussian blur (20\% probability with kernel size 5–9), quality compression (JPEG at 30\%), and shadow addition (brightness factor 0.7) which helped the model to handle various types of image imperfections.

Geometric and Pose Variations: Geometric and Pose Variations made certain that the model had the capacity to identify emotions based on the positioning of the face. Applying rotation up to 30°, translation by 0.1, scaling from 0.8 to 1.2, and flipping with 50\% probability for the horizontal axis and 10\% for the vertical axis were used to improve robustness to face orientations. Pose augmentation, with yaw, pitch, and roll changes, helped to further enhance the model’s accuracy for different head orientations.

Occlusion and Noise Variations: Occlusion and Noise Variations focused on cases where faces could be partially occluded or noisy. Salt and pepper noise was performed with the noise perturbation level set to 5\% while random erasing data augmentation was performed with a probability of 20\% to mimic occlusion by objects in the image. These augmentations allowed the model to be sensitive to recognise gestures or facial expressions in conditions such as noise or occlusions.

\subsection{Model Architecture for classification}
A new and efficient convolutional neural network (CNN) that is specially designed for accurate classification of the facial expressions differentiating between HAPPY and SAD is presented in the proposed design. To enhance feature extraction, and to categorise the features for improved classification, the designed model incorporates Convolutional layers, Batch Normalisation functions, Rectified Linear Unit (ReLU) Nonlinearity, Pooling layers, Dropout regularisation technique ,and Squeeze and Excitation (SE) block, or Convolutional Block Attention Module (CBAM). The proposed architecture is partitioned into two main parts, which are head and body. The head is designed to give out the classification result, while the body is comprised of layers exclusively committed to feature extraction. The pictorial representation of the architecture of our model is depicted below in figure \ref{Figure:10}.

\begin{figure}[H]
\begin{adjustwidth}{-\extralength}{0cm}
\centering
\includegraphics[height=4cm]{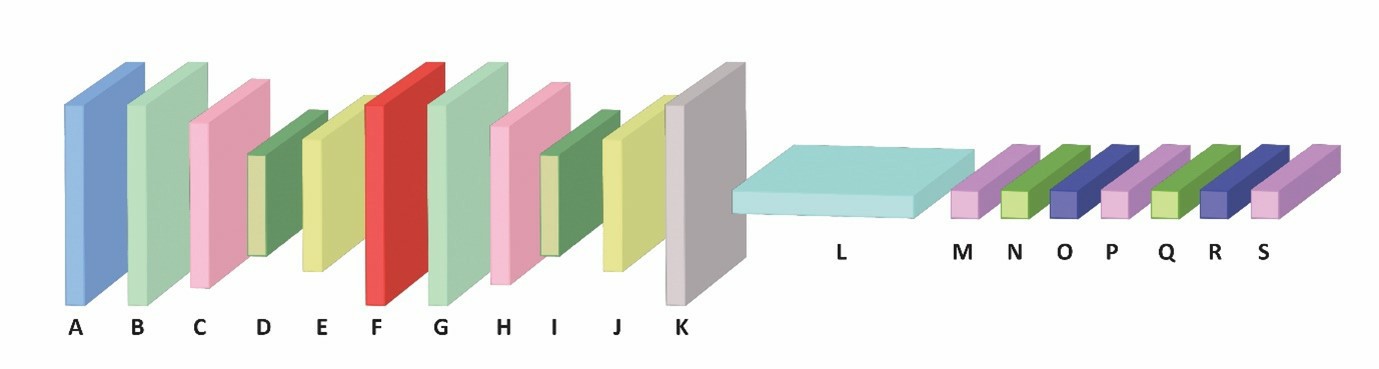}
\end{adjustwidth}
\caption{Our Model Architecture (A: Input; B,G: Conv2D; C,H: Batch Normalization; D,I: ReLU Activation; E,J: Max Pooling; F: SE(Squeeze and Excitation) Block; L: Flattening Layer; M,P,S: Dense Layer; K,O,R: Dropout) }
\label{Figure:10}
\end{figure}

Eight different tests were carried out, each incorporating unique alterations, to assess the efficacy of this architecture:

\begin{itemize}
    \item Experiment 1: base model in which regularisation methods are absent.
    \item Experiment 2: Including a squeeze-and-excitation (SE) block to enhance feature representation by recalibrating channel-wise features adaptively
    \item Experiment 3: Refined feature extraction by the integration of channel and spatial attention mechanisms through the Convolutional Block Attention Module (CBAM).
    \item Experiment 4: By normalising activations, Batch Normalisation (Batch-Norm) is added to stabilise and speed up training.
    \item Experiment 5: Dropout regularisation is used during training to randomly deactivate units in order to prevent overfitting.
    \item Experiment 6: combining dropout and batch normalisation to maximise the advantages of each method.
    \item Experiment 7: To improve feature recalibration while preserving regularisation, Batch Normalisation, dropout, and the SE attention mechanism are integrated.
    \item Experiment 8: Batch Normalisation, dropout, and CBAM attention are implemented in order to investigate the combined effects of these cutting-edge methods.
\end{itemize}

For a comprehensive analysis of the impact of such architectural modifications on the classification of facial expressions, related metrics such as accuracy, confusion matrices, and training/validation loss were applied to each of the experimental trials.

\subsubsection{Feature Extraction}
\paragraph{\textbf{Convolution Layer:}}First, a Convolutional (Conv2d) layer is included in the model which operates on the input picture thereby defining its dimensions from 224*224*3 to 111*111*5 in light of the stated kernel size and pooling techniques. The architecture that we employ in our system is called Conv2d layers, which are designed to learn spatial hierarchies that are inherent to the information extracted from input photos, and do so autonomously. These layers are supposed to prepare essential features needed for the categorisation of the facial expression, such as the shapes, expressions, and textures on the face. After passing through the first Conv2d layer, the dimensions of the image are decreased and, after passing through the second Conv2d layer, the feature representations are increased to provide an output of 54 × 54 × 11.

\paragraph{\textbf{Batch-Norm:}}To increase performance and training stability, the authors use batch normalisation. Because it normalises the activation to zero mean and one variance, it is utilised to stabilise the training process. By doing so, it resolves the Internal Covariate Shift issue, which causes the distribution of activation to shift with each training cycle of the network, slowing convergence and decreasing training stability. Batch Normalisation helps accelerate convergence and raises learning rates by normalising the activations so that all layers see evenly scaled input data. The following formula is used to perform the batch normalisation:

\paragraph{\textbf{Max pooling:}}In order to minimise the spatial dimensions of the feature maps, our model employs max pooling following each convolutional layer in the body. To boost efficiency, it's critical to lessen memory and computational requirements. It reduces the size from 222x222x5 to 111x111x5 to 54x54x11 without sacrificing the most critical features or less relevant information. Additionally, the procedure introduces spatial invariance, which enhances model performance and generalisation.

\subsubsection{Classification}
The classification process in this architecture will be carried out by a sequence of fully connected layers, each of which is intended to convert the extracted data into a final prediction that distinguishes between "happy" and "sAD" facial expressions. The dimensionality continuously decreases as the flattened feature vector is fed into these layers, and ReLU activation functions are employed to capture intricate patterns required for proper classification. After each layer, dropout is performed methodically to enhance generalisation and lower the chance of overfitting. In order to ascertain the anticipated facial expression and guarantee that this model will be successful in differentiating between the two expressions, the last output layer offers a two-dimensional vector that will be the unprocessed class scores.

\subsection{Training Process}
Model parameters are modified during training to maximise classification accuracy. The system processes data in batches and makes use of GPU resources for quicker computations. The cross-entropy loss function analyses the difference between predictions and actual labels from a forward pass and directs changes to minimise it. Stochastic Gradient Descent (SGD) with momentum is used to update parameters after each epoch, accelerating convergence by considering previous updates. As training goes on, a learning rate scheduler progressively lowers the learning rate to enable more precise changes. To keep an eye on performance and direct any required adjustments, key indicators including accuracy and loss are frequently monitored. In order to make sure the model performs well when applied to unknown data, the main objective is to decrease loss while increasing accuracy.

\subsubsection{Hyper-parameter tuning}
For training efficiency and model performance to be maximised, hyperparameter adjustment is essential. Key factors like batch size, learning rate, and epoch count were changed in this work to strike a compromise between computing cost and accuracy. To guarantee steady gradient estimations, a batch size of 32 was selected. An initial learning rate of 0.02 was used, which was lowered by a factor of 0.1 for better fine-tuning in subsequent training phases. In order to ensure quicker convergence and fewer loss oscillations, the optimisation was carried out using Stochastic Gradient Descent (SGD) with a momentum of 0.9. To prevent overfitting and guarantee adequate learning, the model was trained over 40 epochs. Ten employees and a GPU were also used for effective data loading. Based on experimental findings, Table 4 provides a summary of all the hyperparameters.

\begin{table}[h!]
\centering
\caption{Details of Hyperparameters Utilized in the Training Process}
\centering
\begin{tabular}{|>{\centering\arraybackslash}p{5cm}|>{\centering\arraybackslash}p{5cm}|}
\hline
\textbf{Hyperparameter} & \textbf{Value} \\ \hline
Batch Size & 32 \\ \hline
Number of Epochs & 40 \\ \hline
Initial Learning Rate & 0.02 \\ \hline
Learning Rate Decay Rate & 0.1 \\ \hline
Optimizer & SGD \\ \hline
Momentum & 0.9 \\ \hline
Number of Workers & 10 \\ \hline
\end{tabular}
\label{tab:hyperparameters}
\end{table}

\section{Evaluation Matrix}
Therefore, several measurements that include accuracy, precision, recall rate, and F1 score can be employed in measurement of the model’s outcome. All of these aspects help the model to differentiate between different face emotions, say HAPPY and SAD.

\subsection{Accuracy}
This measure gives the ratio of all correctly identified instances over the total number of instances considered in the analysis. It is defined solely by the formula whereby it is arrived at, by dividing the number of correct forecasts by the total number of forecasts made. The accuracy formula is as follows:

\begin{equation}
    \text{Accuracy} = \frac{\text{True Positive} + \text{True Negetive}}{\text{True Positive} + \text{True Negetive} + \text{False Positive} + \text{False Negetive}}
\end{equation}

Although it provides a broad sense of the models' effectiveness, this metric is not always indicative of that effectiveness, especially when there is a class imbalance.

\subsection{Precision}
Precision is the measure of the model’s positive predictions that are correct to a large extent.  It is the ratio of true positive predictions to expected positive predictions which includes false positive predictions also. The source of precision is:

\begin{equation}
    \text{Precision} = \frac{\text{TP}}{\text{TP} + \text{FP}}
\end{equation}

In order to determine the credibility of positive predictions, precision is essential when the cost of false positives is significant.

\subsection{Recall}
recall, also called Sensitivity, quantifies how well a machine learning algorithm can identify all the positive instances. In this context, it defines the proportion of true positives to the overall number of actual positive cases which equal those that were categorised as true positive and false negative. The formula for recall is:

\begin{equation}
    \text{Recall} = \frac{\text{TP}}{\text{TP} + \text{FN}}
\end{equation}

Recall becomes a key performance metric, especially in highly expensive situationswhere it is impermissible to miss any positivity cases as it reveals the efficiency achieved at positive identification.

\subsection{F1-score}
In this context, the F1-score offers an objective and fair measurement for the performance of ML algorithms by considering both the precision and recall rates. This is particularly helpful in the case of datasets with significant class imbalance. This is how the F1-score is computed:

\begin{equation}
    F_1 \text{-score} = \frac{2 \cdot \text{Precision} \cdot \text{Recall}}{\text{Precision} + \text{Recall}}
\end{equation}

In this way, the F1-score can demonstrate the accuracy of positive predictions while also comparing them to all pertinent positive examples. That is why it is employed in circumstances where an effective performance evaluation requires a balance between precision and recall.


\begin{table}[h!]
\centering
\caption{Performance metrics of different models and configurations for Happy and Sad emotion detection.}
\resizebox{\textwidth}{!}{%
\begin{tabular}{|l|ccc|ccc|c|}
\hline
{\textbf{Model}} & \multicolumn{3}{c|}{\textbf{Happy}} & \multicolumn{3}{c|}{\textbf{Sad}} & \textbf{Overall} \\
                                 & \textbf{Precision} & \textbf{Recall} & \textbf{F1-Score} & \textbf{Precision} & \textbf{Recall} & \textbf{F1-Score} & \textbf{Accuracy} \\ \hline
Model 1: Small Custom Model [S] & & & & & & & \\ \hline
Experiment 1: Without Regularization [S] & 0.81 & 0.77 & 0.79 & 0.78 & 0.82 & 0.80 & 0.80 \\ \hline
Experiment 2: With Attention Block (SE Block) & 0.87 & 0.82 & 0.85 & 0.83 & 0.88 & 0.85 & 0.85 \\ \hline
Experiment 3: With CBAM (Channel + Spatial Attention) & 0.84 & 0.83 & 0.84 & 0.83 & 0.84 & 0.84 & 0.84 \\ \hline
Experiment 4: With BatchNorm [M] & 0.88 & 0.83 & 0.86 & 0.84 & 0.89 & 0.87 & 0.86 \\ \hline
Experiment 5: With Dropout [M] & 0.85 & 0.89 & 0.87 & 0.89 & 0.84 & 0.86 & 0.87 \\ \hline
Experiment 6: With BatchNorm and Dropout [M] & 0.90 & 0.87 & 0.89 & 0.87 & 0.90 & 0.87 & 0.88 \\ \hline
Experiment 7: With BatchNorm, Dropout, and SE Attention & 0.91 & 0.86 & 0.88 & 0.87 & 0.92 & 0.89 & 0.89 \\ \hline
Experiment 8: With BatchNorm, Dropout, and CBAM Attention & 0.91 & 0.80 & 0.85 & 0.82 & 0.92 & 0.87 & 0.86 \\ \hline
\end{tabular}%
}
\label{tab:performance}
\end{table}

\section{Discussion}
\subsection{Comparative anaysis of experiments}
The effects of various model configurations on performance are displayed in Table 5. Without regularisation, the accuracy of the baseline model was 0.80. Accuracy improved to 0.85 after the Squeeze-and-Excitation (SE) Block was included, strengthening the model's focus on important characteristics and boosting precision and recall for both expression categories. Performance was further improved by adding the Convolutional Block Attention Module (CBAM), which achieved an accuracy of 0.84 but fell short of the gains made with the SE Block. By decreasing overfitting, Dropout improved the findings to 0.87, while Batch Normalisation increased the accuracy to 0.86. Combining Batch Normalisation, Dropout, and SE Attention the model produced the optimal results, with an accuracy of 0.89, indicating that the combination well balanced F1 scores, precision, and recall. However, even with such a high number of feature maps, it was impossible to achieve better results than the SE Attention configuration for Batch Normalisation, Dropout, and CBAM. Such results give the general idea of how effectively the advanced methods can enhance facial emotion recognition efficiency.

\subsection{Comparison with SOTA models}
It can be observed in Table 5 that our proposed model inculcating batch regularisation, Dropout, Squeezed and SE attention blocks yielded a single accuracy of 0.89. Although achieving the best result in facial expression classification, this result is not S.O.T.A. models that have been achieved. For instance, MobileNet is regarded as having a high accuracy of 0.99, meaning that with its high accuracy and speed, it can be considered to be among the most efficient AI models. Similarly, ResNet18 has achieved 0.98 and GoogLeNet achieved 0.961 showing high efficiency of their deep learning architecture. In contrast, AlexNet’s performance seems rather poor compared to these more recent networks, achieving only a 0.5 accuracy. This comparison demonstrates that while our model does well, SOTA models like MobileNet, ResNet18, and GoogLeNet offer better accuracy, demonstrating their more advanced and efficient design in facial expression classification.

\begin{table}[h!]
\centering
\caption{Comparative Analysis of Model Performance against SOTA Models}
\label{table:comparative_analysis}
\begin{tabular}{|>{\centering\arraybackslash}p{10cm}|>{\centering\arraybackslash}p{3cm}|}
\hline
\textbf{Models} & \textbf{Accuracy} \\ \hline
ResNet18 & 0.98 \\ \hline
AlexNet & 0.5 \\ \hline
GoogLeNet & 0.961 \\ \hline
MobileNet & 0.99 \\ \hline
Our Proposed Model (With BatchNorm, Dropout, and SE Attention) & 0.89 \\ \hline
\end{tabular}
\end{table}

\section{Ethical and Legal Considerations}
In developing the facial expression classification model for children, there are legal and ethical consideration followed strictly to ensure that the objectives set are met. The parents or guardians’ consent is not required due to the usage of solely publicly available images and around 90\% of the dataset are the synthetic images generated by advanced AI techniques. Methods such as anonymisation and robust measures ensure data protection and the conception of the study is intended to cause as little distress as possible to the participants. No unauthorised commercial use ensures compliance with legalities, for example, the Children’s Online Privacy Protection Act (COPPA) and the General Data Protection Regulation (GDPR). All the procedures are conducted in line with the prevailing laws and ethics.

\section{Conclusion}
Compared to adult-oriented platforms, developing a specialised model for categorising children's facial expressions fills a significant gap by improving emotion recognition by concentrating on how kids convey the "Happy" and "Sad" emotions. Along with robust data augmentation, the model's performance has been enhanced by sophisticated techniques, especially attention mechanisms like Squeeze-and-Excitation blocks and Convolutional Block Attention Modules. This was made possible through stable diffusion which led to the diversification and expansion of the dataset as well as the development of more realistic training samples.

Batch Normalisation with both Dropout and SE Attention achieved the highest accuracy, which proves just how effective these techniques are in improving generalisation. Thus, to follow the model’s potential, additional advancements are still needed to provide the same or higher accuracy than models like MobileNet and ResNet18, extend the range of classification emotions, reduce the frequency of misunderstandings, and diversify and enhance the developed synthesised images. Future studies into more complex frameworks and training methods will be essential to tackling practical issues and creating a universal tool to safeguard kids' emotional health on the internet. Additionally, our work can be extended to other domains such as renewable energy ~\cite{hussain2022statistical} where data manipulation is required before training models ~\cite{hussain2019deployment}.

\begin{adjustwidth}{-\extralength}{0cm}

\bibliographystyle{unsrt}  
\bibliography{my_references}  

\end{adjustwidth}
\end{document}